\pdfoutput=1

\documentclass{svmult}

\usepackage[utf8]{inputenc}
\usepackage{sectsty}
\usepackage[table]{xcolor}  
\usepackage{mathptmx}       
\usepackage{helvet}         
\usepackage{courier}        
\usepackage{type1cm}        
\usepackage{makeidx}         
\usepackage{graphicx}        
\usepackage{multicol}        
\usepackage[bottom]{footmisc}
\usepackage{amsmath,amssymb,epsfig}

\usepackage{epstopdf,overpic}

\makeindex             

\begin{document}

\title*{Rational Parametrization of Linear Pentapod's Singularity Variety and the Distance to it}
\author{Arvin Rasoulzadeh and Georg Nawratil}
\authorrunning{A. Rasoulzadeh and G. Nawratil}
\institute{Center for Geometry and Computational Design, Vienna University of Technology, Austria, 
\email{\{rasoulzadeh, nawratil\}@geometrie.tuwien.ac.at} 
}

%
%
%
\maketitle

\abstract{A linear pentapod is a parallel manipulator with five collinear anchor points on the motion platform (end-effector), 
which are connected via $\mathrm{S\underline{P}S}$ legs to the base. This manipulator has five controllable degrees-of-freedom and the remaining one is a free rotation around the motion platform axis (which in fact is an axial spindle). 
In this paper we present a rational parametrization of the singularity variety of the linear pentapod. 
Moreover we compute the shortest distance to this rational variety with respect to a suitable metric. 
Kinematically this distance can be interpreted as the radius of the maximal singularity free-sphere. 
Moreover we compare the result with the radius of the maximal singularity free-sphere in the position workspace and the 
orientation workspace, respectively.     
} 

\section{Introduction}\label{sec:intro}


The Stewart-Gough platform (sometimes called simply Stewart platform) can be defined as six degree-of-freedom (DOF) 
parallel manipulator (PM) with six identical spherical-prismatic-spherical ($\mathrm{S}\underline{\mathrm{P}}\mathrm{S}$) 
legs, where only the prismatic joints are active.  
This parallel robot is merely used in \emph{flight simulation} where a replica cockpit plays the role of the moving platform. 

\begin{figure}[h!] 
\begin{center}   
  \begin{overpic}[height=71mm]{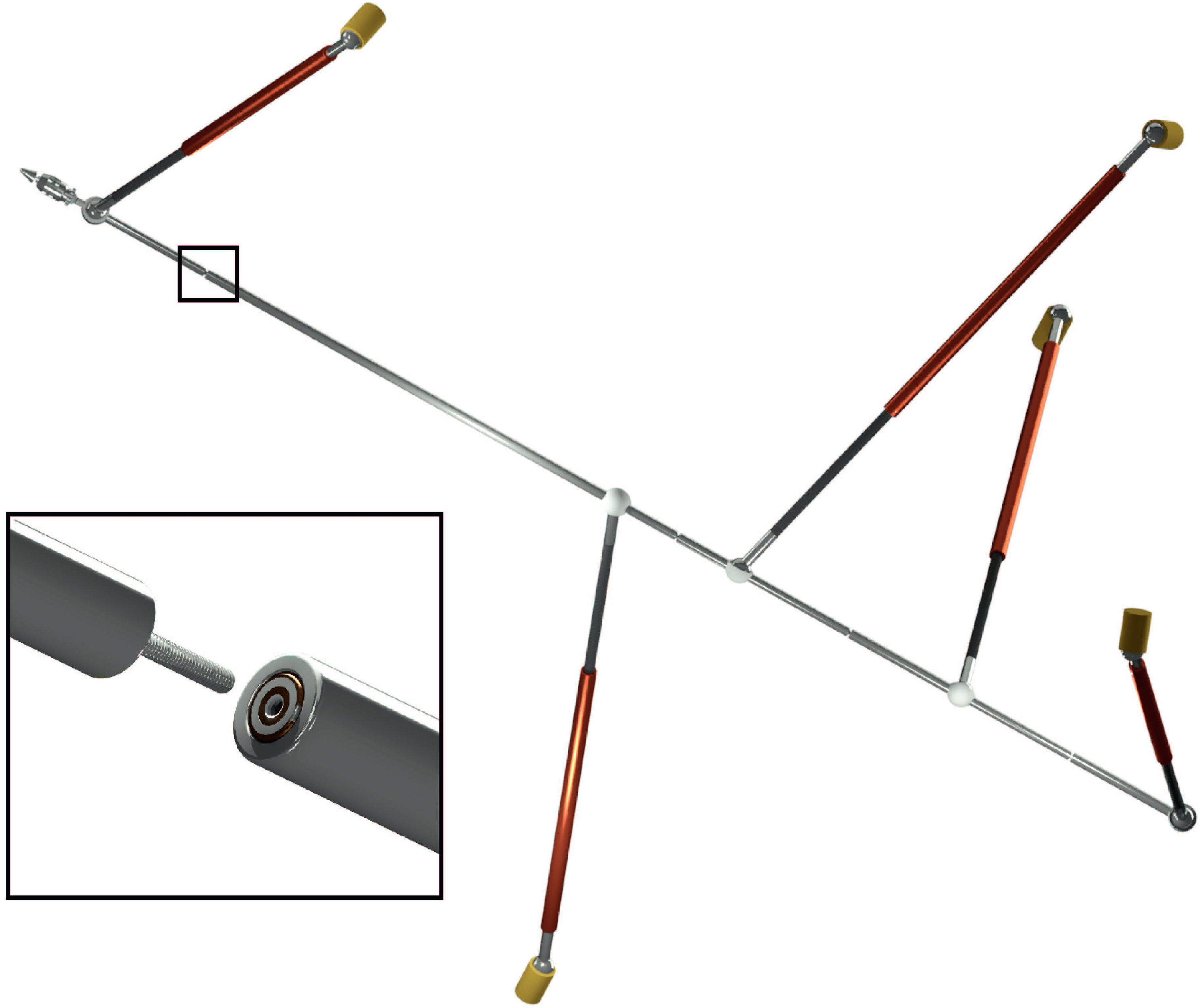}
	\put(34.5,53){$\ell$}
  \end{overpic} 
	\caption{Linear pentapod with the following architectural parameters: 
	\leavevmode\ \ 
	 $\mathbf{a}_{1}=(0,0,0)^{T}$, $\mathbf{a}_{2}=(5,0,0)^{T}$, $\mathbf{a}_{3}      =(-4,-3,0)^{T}$, $\mathbf{a}_{4}=(3,7,-6)^{T}$, $\mathbf{a}_{5}=(9,-5,4)^{T}$, $(r_{1}, r_{2}, r_{3}, r_{4}, r_{5})=(0,2,4,5,10)$. 
	Moreover it should be noted that in the  illustrated design the linear platform $\ell$ consists of five parts, which are 
	jointed by four passive rotational joints (a zoom of this detail is given in the box). 
	This construction enlarges the workspace by compensating some joint limits of the platform S-joints.}
	\label{fig:1}
\end{center}
\end{figure}

Although the Stewart platform is the most celebrated PM, some of its sub-assemblies with a lower number of legs are of interest 
from theoretical and practical points of view. Sometimes these sub-assemblies are referred to as \emph{components} \cite{kg2000}. 
In this paper we study the so-called \emph{line-body component}, which is  
a rigid sub-assembly of a Stewart PM consisting of a linear motion platform (end-effector) named $\ell$ and 
five  $\mathrm{S}\underline{\mathrm{P}}\mathrm{S}$ legs, where the base anchor points can have position in $\mathbb{R}^{3}$. 
Here this component is referred to as \emph{linear pentapod}, which is
an alternative to serial robots for handling axis-symmetric tools (see Fig.\ \ref{fig:1}). Moreover we use the 
following notations: 
\begin{enumerate}
\item[1.]The position of $\ell$ is given by the vector $\mathbf{p}=(p_{x}, p_{y}, p_{z})^{T}$ and the orientation 
of $\ell$ is defined by a unit-vector $\mathbf{i}=(u, v, w)^{T}$.
\end{enumerate}  
\begin{enumerate}
\item[2.]The coordinate vector $\mathbf{b}_{j}$ of the platform anchor point of the $j$th leg is described by the 
equation $\mathbf{b}_{j}=\mathbf{p}+r_{j}\mathbf{i}$ for $j=1,\ldots ,5$.
\item[3.]The base anchor point of the leg $j$ has coordinates $\mathbf{a}_{j}=(x_{j}, y_{j}, z_{j})^{T}$.
\end{enumerate}  
Note that all vectors are given with respect to a fixed reference 
frame, which can always 
be chosen and scaled in a way that the following conditions hold:
\begin{equation}\label{assumptions}
x_{1}=
y_{1}=
z_{1}= 
y_{2}=
z_{2}=
z_{3}=0 \quad \text{and}\quad x_{2}=1.
\end{equation}

According to \cite[Theorem 12]{naccepted} one possible point-model for the configuration space $\mathcal{C}$ of the linear 
pentapod reads as follows: {\it There exists a bijection between $\mathcal{C}$ and all real points 
$\mathfrak{C}=(u,v,w,p_x,p_y,p_z)\in\mathbb{R}^{6}$ located on the singular quadric $\Gamma:\,\,u^2+v^2+w^2=1$.} 
Based on this notation we study the singularity loci of linear pentapods and the distance to it in the paper at 
hand, which is structured as follows:




We close Section \ref{sec:intro} by a review on the singularity analysis of linear pentapods and 
recall the implicit equation of the singularity variety.  
In Section \ref{sec:rat} we give a brief introduction to rational varieties and
present a rational parametrization of the singularity loci of linear pentapods. 
In Section \ref{sec:dis} we compute the minimal distance to the singularity variety with respect to a
novel metric in the ambient space $\mathbb{R}^6$ of the configuration space $\mathcal{C}$. 
We also compute the closest singular configuration under the constraint of a fixed 
orientation and a fixed position, respectively. 
Finally a conclusion and a plan for future research is given.

\subsection{Singularity Variety of the Pentapod}

Singularity analysis plays an important role in motion planning of PMs.  
For linear pentapods the singularities as well as the singular-invariant leg-rearrangements 
have been studied in  \cite{btt2011} 
for a planar base and in \cite{borras2010singularity} for a non-planar one. 
A complete list of architectural singular designs of linear pentapods is given in \cite{ns},  
where also non-architecturally singular designs with self-motions are classified (see also \cite{n2016}).

Kinematical singularities occur whenever the Jacobian matrix $\mathbf{J}$ becomes rank deficient, 
where $\mathbf{J}$ can be written as follows (cf.\ \cite{borras2010singularity}):
\begin{equation*}
\mathbf{J}=
\begin{pmatrix}
\mathbf{l}_{1} & \dots & \mathbf{l}_{5} \\
\mathbf{\hat l}_{1} & \dots & \mathbf{\hat l}_{5}
\end{pmatrix}^T
\text{with}\,\,\,
\mathbf{l}_{j}=
\begin{pmatrix}
p_{x}+r_{j}u-x_{j}\\
p_{y}+r_{j}v-y_{j}\\
p_{z}+r_{j}w-z_{j}
\end{pmatrix},\,\, 
\mathbf{\hat l}_{j}=
\begin{pmatrix}
z_{j}(p_{y}+r_{j}v)-y_{j}(p_{z}+r_{j}w)\\
x_{j}(p_{z}+r_{j}w)-z_{j}(p_{x}+r_{j}u)\\
y_{j}(p_{x}+r_{j}u)-x_{j}(p_{y}+r_{j}v)
\end{pmatrix}.
\end{equation*} 
This $5\times 6$ Jacobian matrix $\mathbf{J}$ has a rank less than five whenever the determinants of all its $5\times 5$ sub-matrices vanish. So by naming the determinant of the $5\times 5$ sub-matrix, which results from excluding the $j$th column, with $F_{j}$ 
the singularity loci equals $V(F_{1},\dots , F_{6})$; i.e.\ the variety of the ideal spanned by the polynomials $F_{1},\dots , F_{6}$. 
It can easily be checked by direct computations that this variety equals the zero-set of the greatest common divisor $F$ of 
$F_{1},\dots , F_{6}$. This singularity polynomial $F$ has the following structure:
\begin{equation}\label{singularity}
\begin{split}
F:=
&(A_{{1}}p_{{y}}+A_{{2}}p_{{z}}) {u}^{2}+[(A_{{3}}p_{{x}}+A_{{4}}p_{{y}}+A_{{5}}p_{{z}}+A_{{6}}) v+( A_{{7}}p_{{x}}+A_{{8}}p_{{y}}
\\
&+A_{{9}}p_{{z}}+A_{{10}})w + (A_{{11}}p_{{y}}+A_{{12}}p_{{z}})p_{{x}}+A_{{13}}{p_{{y}}}^{2}+ (A_{{14}}p_{{z}}+A_{{15}})p_{{y}}
\\
&+A_{{16}}{p_{{z}}}^{2}+A_{{17}}p_{{z}}]u+(A_{{18}}p_{{x}}+A_{{19}}p_{{z}}+A_{{20}}){v}^{2}+[( A_{{21}}p_{{x}}+A_{{22}}p_{{y}}
\\
&+A_{{23}}p_{{z}}+A_{{24}})w+A_{{25}}{p_{{x}}}^{2}+(A_{{26}}p_{{y}}+A_{{27}}p_{{z}}+A_{{28}})p_{{x}}+( A_{{29}}p_{{z}}
\\
&+A_{{30}})p_{{y}}+A_{{31}}{p_{{z}}}^{2}+A_{{32}}p_{{z}}]v+( A_{{33}}p_{{x}}+A_{{34}}p_{{y}}+A_{{35}}) {w}^{2}+[ A_{{36}}{p_{{x}}}^{2}
\\
&+ ( A_{{37}}p_{{y}}+A_{{38}}p_{{z}}+A_{{39}}) p_{{x}}+A_{{40}}{p_{{y}}}^{2}+(A_{{41}}p_{{z}}+A_{{42}})p_{{y}}+A_{{43}}p_{{z}}]w
\end{split}
\end{equation}
where the coefficients $A_{i}$ belong to the ring $\mathrm{R}=\mathbb{R}[x_{3},x_4, x_{5},y_{3},y_4 ,y_{5},z_{4} ,z_{5},r_{1},\ldots ,r_{5}]$  
which evidently makes $F$ a polynomial with the total-degree of 3 belonging to $\mathrm{R}[u, v, w, p_{x}, p_{y}, p_{z}]$. 
Note that for a specified orientation $(u,v,w)$ the equation $F=0$ 
determines only a quadric surface $\Omega(u,v,w)$ in the space of positions. 
This property is of great importance later on.

\begin{remark}
It can easily be checked that the polynomial $F$ is identical with the determinant of a $7\times 7$ matrix, 
given in \cite[Eq.\ (4)]{borras2010singularity}. 
\hfill $\diamond$
\end{remark}

\section{Rational Parametrization of the Singularity Variety}\label{sec:rat}

In this section we rationally parametrize the singularity variety, which is given by 
the implicit equation $F=0$. But before stepping into the computations, the presentation of a 
formal definition of this parametrization seems necessary. 
\begin{definition}\label{def1}
\label{definition1}
Let $\mathbb{K}$ be a field and $V\subset\mathbb{K}^{m}$ and $W\subset\mathbb{K}^{n}$ be irreducible affine varieties. 
A \emph{rational mapping} from $V$ to $W$ is a function $\phi$ represented by
\begin{equation}
\phi:V \dashrightarrow W \quad \text{with}\quad
\phi (x_{1}, \dots , x_{m})=\left(\frac{f_{1}(x_{1}, \dots ,x_{m})}{g_{1}(x_{1}, \dots ,x_{m})},\dots 
, \frac{f_{n}(x_{1}, \dots ,x_{m})}{g_{n}(x_{1}, \dots ,x_{m})}\right)
\label{para}
\end{equation}  
where $\frac{f_{i}}{g_{i}}\in \mathbb{K}(x_{1},\dots ,x_{m})$ and satisfies the following properties:
\begin{enumerate}
\item[1.]$\phi$ is defined at some point of $V$.
\item[2.]For every $(a_{1},\dots a_{m})\in V$ where $\phi$ is defined, $\phi (a_{1},\dots a_{m})\in W$. 
\end{enumerate}
\end{definition}
\begin{definition}\label{def2}
\label{definition2}
Two irreducible varieties $V$ and $W$ are said to be \emph{birationally equivalent} if there exist rational mappings $\phi :V \dashrightarrow W$ and $\psi :W \dashrightarrow V$ such that $\phi \circ \psi$ and $\psi \circ \phi$ be equal to $\mathrm{id}_{W}$ and $\mathrm{id}_{V}$ respectively.
\end{definition} 
\begin{definition}\label{def3}
\label{definition3}
A \emph{rational variety} is a variety that is \emph{birationally} equivalent to $\mathbb{K}^{n}$.
\end{definition}
One can find the extensive discussion of above definitions in \cite[Chapters 1 and 2]{shafarevich1977basic}. 

Having a rational parametrization of a variety has numerous advantages: 
If the coefficients of the polynomials $f_{i}$ and $g_{i}$ of Eq.\ (\ref{para}) belong to $\mathbb{Q}$ and if  $(x_1,\ldots ,x_m)$  
is an element of $\mathbb{Q}^{m}$, then one obtains points with rational coordinates on the singularity variety \cite[page 3]{shafarevich1977basic}. 
This is a matter, which is of high importance to computer aided designs, as computers can calculate rational coordinates 
at a much faster rate. 


Moreover the rationality of the singularity variety implies that it is \emph{path connected}, 
which means that every singular pose can be connected to any other singular pose by a continuous singular 
motion \cite{husty2008singularity}. This property can be used for a computationally efficient approximation 
of the singularity-free workspace by hierarchical structured hyperboxes, where only their boundaries have to be checked to be 
free of singularities. 
Beside the rationally parametrized singularity loci of the planar  3-R\underline{P}R PM \cite{husty2008singularity}, 
only the one of Stewart PMs with planar platform and planar base \cite{cm2015} (see also \cite{an2016,bg2006}) 
are known to the authors (in the context of PMs of Stewart-Gough type). 

For the computation of the rational parametrization of the linear pentapod, we exploit the idea used in \cite{cm2015}: 
By homogenizing the singularity polynomial $F$ of Eq.\ (\ref{singularity}) by the extra variable $p_{0}$ with respect 
to the position variables $p_{x}$, $p_{y}$ and $p_{z}$, we obtain a homogeneous polynomial 
$F_{h}\in\mathrm{R}(u,v,w)[p_{x},p_{y},p_{z},p_{0}]$ in the projective 3-space $\mathbb{P}^{3}$ with 
homogeneous  coordinates $(p_{x}:p_{y}:p_{z}:p_{0})$.  
It turns out that the point $\sf B$ with homogeneous coordinates $(u: v: w: 0)$ 
is a point of the singularity variety; i.e. $\sf B\in V(F_{h})\subset \mathbb{P}^{3} $. 
Note that $\sf B$ is the ideal point of the linear platform $\ell$ with orientation vector $\mathbf{i}$. 

The side condition on the vector $\mathbf{i}=(u,v,w)^T$ to be of unit-length, can be avoided by 
using the \emph{stereographic parametrization} of the unit-sphere $S^{2}$: 
\begin{equation}
\mathbf{x}:(t_{3},t_{4})\mapsto \left(\frac{2\ t_{3}}{{t_{3}}^{2}+{t_{4}}^{2}+1}, 
\frac{2\ t_{4}}{{t_{3}}^{2}+{t_{4}}^{2}+1}, \frac{{t_{3}}^{2}+{t_{4}}^{2}-1}{{t_{3}}^{2}+{t_{4}}^{2}+1}\right).
\end{equation}
Based on this we can parametrize the lines of the bundle $\mathcal{B}$ with vertex $\sf B$ 
in the finite space $\mathbb{R}^{3}$ of positions with coordinates $(p_{x}, p_{y},p_{z})$ as follows:  
\begin{equation}
\mathcal{B}: \left(\begin{array}{c} p_{x} \\ p_{y} \\ p_{z}  \end{array}\right)= a\mathbf{x}(t_{3},t_{4})+
t_{1}\frac{\partial\mathbf{x}(t_{3},t_{4})}{\partial t_{3}}+t_{2}\frac{\partial\mathbf{x}(t_{3},t_{4})}{\partial t_{4}}.
\label{equ}
\end{equation}
Note that the bituple $(t_1,t_2)$ fixes the line of the bundle $\mathcal{B}$ and the parameter $a$ 
determines the point on this line. By varying $(t_1,t_2)\in\mathbb{R}^{2}$ and setting $a=0$ one 
obtains the plane through the origin, which is orthogonal to $\mathbf{i}$. 

Plugging $\mathcal{B}(a,t_1,t_2,t_3,t_4)$ into 
$F=0$ 
shows that 
the resulting expression is only linear in $a$, as the ideal point ${\sf B}$ is always one of the two intersection points 
of a line belonging to $\mathcal{B}$ with the quadric $\Omega(\mathbf{x}(t_3,t_4))$. By solving this linear condition we get 
$a(t_{1},t_{2},t_{3}, t_{4})$. 
Now the singular configurations $\mathfrak{X}=(\xi_1,\ldots ,\xi_6)\in\mathbb{R}^6$ of the linear pentapod can be rationally parametrized 
by $(\xi_{1},\xi_{2},\xi_{3}):=\mathbf{x}(t_3,t_4)$ and  
\begin{equation}
\begin{split}
%
%
%
\xi_{4}& =2\,{\frac {a \left( t_{{1}},t_{{2}},t_{{3}},t_{{4}} \right) t_{{3}}}{{
t_{{3}}}^{2}+{t_{{4}}}^{2}+1}}- 2\,{\frac {t_{{1}} \left( {t_{{3}}}^{2}
-{t_{{4}}}^{2}-1 \right) }{ \left( {t_{{3}}}^{2}+{t_{{4}}}^{2}+1
 \right) ^{2}}}-4\,{\frac {t_{{2}}\ t_{{3}}\ t_{{4}}}{ \left( {t_{{3}}}^{2
}+{t_{{4}}}^{2}+1 \right) ^{2}}}
,\\
\xi_{5}& =2\,{\frac {a \left( t_{{1}},t_{{2}},t_{{3}},t_{{4}} \right) t_{{4}}}{{
t_{{3}}}^{2}+{t_{{4}}}^{2}+1}}-4\,{\frac {t_{{1}}\ t_{{3}}\ t_{{4}}}{
 \left( {t_{{3}}}^{2}+{t_{{4}}}^{2}+1 \right) ^{2}}}+2\,{\frac {t_{{2}}
 \left( {t_{{3}}}^{2}-{t_{{4}}}^{2}+1 \right) }{ \left( {t_{{3}}
}^{2}+{t_{{4}}}^{2}+1 \right) ^{2}}}
,\\
\xi_{6}& ={\frac {a \left( t_{{1}},t_{{2}},t_{{3}},t_{{4}} \right)  \left( {t_{{
3}}}^{2}+{t_{{4}}}^{2}-1 \right) }{{t_{{3}}}^{2}+{t_{{4}}}^{2}+1}}+4\,
{\frac {t_{{1}}t_{{3}}}{ \left( {t_{{3}}}^{2}+{t_{{4}}}^{2}+1 \right) 
^{2}}}+4\,{\frac {t_{{2}}t_{{4}}}{ \left( {t_{{3}}}^{2}+{t_{{4}}}^{2}+
1 \right) ^{2}}}
.
\end{split}
\end{equation}
This parametrization covers the singular variety with exception of two low-dim\-ensional sub-variety:  
A missing 3-dimensional sub-variety is defined by the denominator of $a(t_{1}, t_{2}, t_{3}, t_{4})$. 
In this case the residual intersection point $\in \mathbb{R}^{3}$ of the line belonging to $\mathcal{B}$ with $\Omega(\mathbf{x}(t_3,t_4))$ 
is not determined uniquely; i.e.\ the complete line belongs to $\Omega(\mathbf{x}(t_3,t_4))$. 
As the orientation $(0,0,1)$ cannot be obtained by the \emph{stereographic parametrization}, also 
the 2-dimensional sub-variety  $\Omega(0,0,1)$ is missing. 

Moreover for a given singular pose $\mathfrak{X}=(\xi_1,\ldots ,\xi_6)\in\mathbb{R}^6$ we 
can trivially compute  $t_1,\ldots, t_4$  in a rational way from $\xi_1,\ldots ,\xi_6$, 
thus the singularity variety is a rational one (according to the Definitions \ref{def1}, \ref{def2} and  \ref{def3}).

\section{Distance to the Singularity Variety}\label{sec:dis}

In singularities the number of DOFs of the mechanism changes instantaneously and becomes uncontrollable. 
Additionally the actuator forces can become very large and cause the break down of the platform \cite{li2007determination}. 
Henceforth knowing the distance of a given pose $\mathfrak{G}=(g_1,\ldots ,g_6)\in\mathbb{R}^6$ 
from the singularity variety is of great importance. 

\begin{figure}[t!] 
\begin{center}   
  \begin{overpic}[height=45mm]{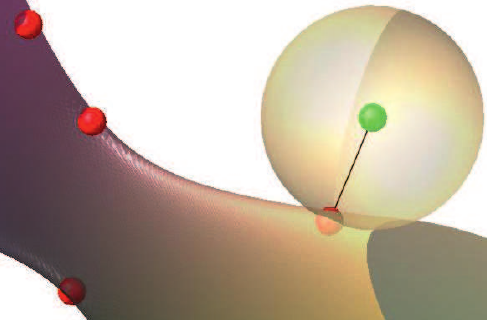}
	\put(36,31){$\Omega$}
  \end{overpic} 
	\hfill
	\begin{overpic}[height=45mm]{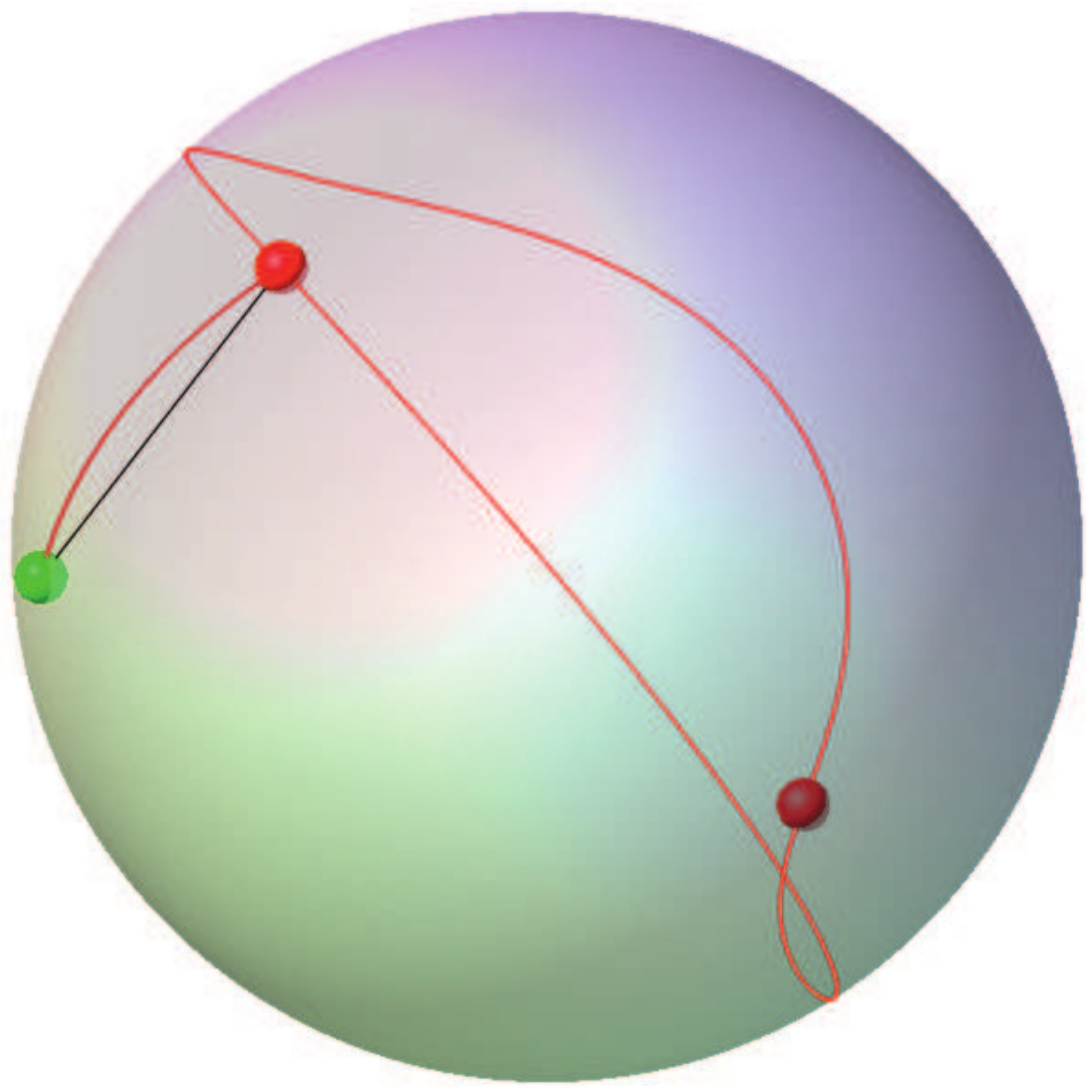}
	\put(34.5,53){$\omega$}
  \end{overpic} 
	\caption{Illustrations are done for $\mathfrak{G}=(\frac{3}{5},\frac{4}{5},0,2,3,4)$ of the linear pentapod displayed in 
	Fig.\ \ref{fig:1}.
	Fixed orientation (Left):  $\mathcal{O}$ has only four real solutions where the closest one 
	$\mathfrak{O}=(\frac{3}{5},\frac{4}{5},0,2.5517,2.6374,0.1144)$ has a distance of 3.9412 units. 
	Fixed position (right): $\mathcal{P}$ has only two real solutions where the closest one 
	$\mathfrak{P}=(0.3701,0.5523,0.7468,2,3,4)$ has a spherical distance of $48.4178^{\circ}$.}
	\label{fig:2}
\end{center}
\end{figure}
\begin{figure}[t!] 
\begin{center}   
  \begin{overpic}[height=51mm]{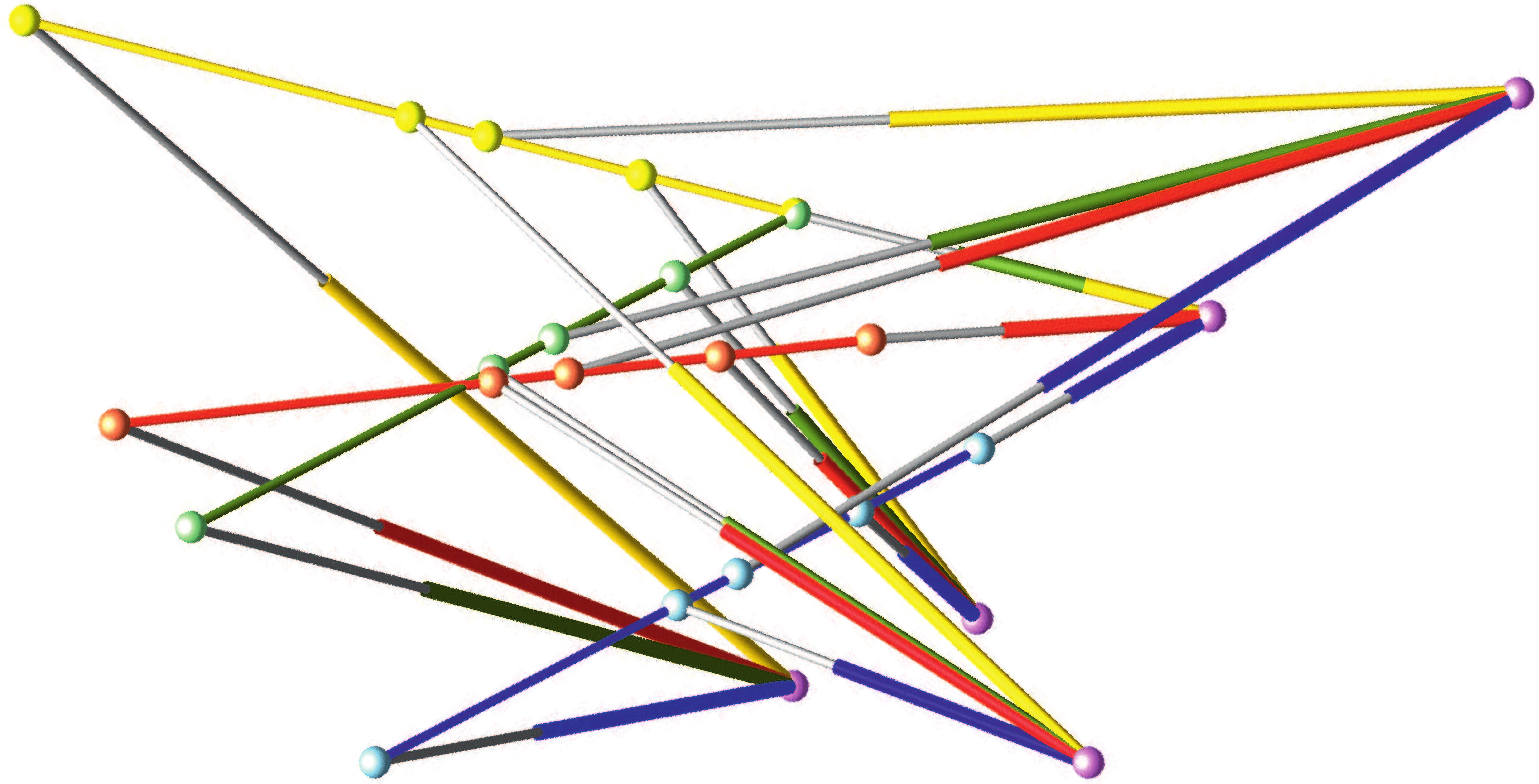}
	\put(14,43){$\mathfrak{P}$}
	\put(15,25.5){$\mathfrak{M}$}
	\put(24,20){$\mathfrak{G}$}
	\put(29,5.5){$\mathfrak{O}$}
  \end{overpic} 
	\caption{Comparison of the different configurations $\mathfrak{G}$ (green), 
		$\mathfrak{O}$ (blue), $\mathfrak{P}$ (yellow) and the red-colored  $\mathfrak{M}=(0.5559, 0.7274, 0.4021, 2.2966, 3.4794, 1.8357)$  
	with $d(\mathfrak{M}, \mathfrak{G})=1.4791$. In contrast $d(\mathfrak{O}, \mathfrak{G})=3.9412$ and $d(\mathfrak{P}, \mathfrak{G})=4.4142$. For this example only 16 of 80 pedal-points are real. 
	}
	\label{fig:3}
\end{center}
\end{figure}
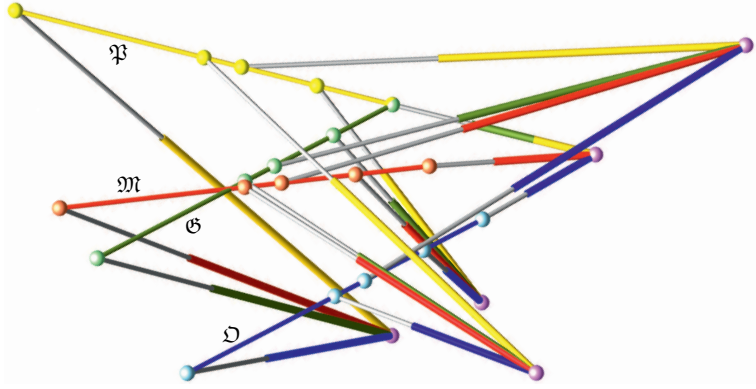

\paragraph{Fixed Orientation:}
We ask for the closest singular configuration $\mathfrak{O}$ having the same orientation $(g_1,g_2,g_3)$ as the given pose $\mathfrak{G}$. 
As  $\mathfrak{G}$ and  $\mathfrak{O}$ only differ by a translation, we can define the distance between 
these two poses by the length of the translation vector. Therefore $\mathfrak{O}$ has to be a pedal-point on $\Omega(g_1,g_2,g_3)$ with 
respect to the point $(g_4,g_5,g_6)$. The set $\mathcal{O}$ of all these pedal-points equals the variety
$V(\tfrac{\partial L}{\partial p_{x}}, \tfrac{\partial L}{\partial p_{y}}, \tfrac{\partial L}{\partial p_{z}}, 
\tfrac{\partial L}{\partial \lambda})$
where $\lambda$ is the Lagrange multiplier of the Lagrange equation
\begin{equation}
L(p_{x}, p_{y},p_{z},\lambda)=(p_{x}-g_{4})^{2}+(p_{y}-g_{5})^{2}+(p_{z}-g_{6})^{2}+\lambda F. 
\end{equation}
It is well known (see Appendix A) that in general $\mathcal{O}$ consists of six points over $\mathbb{C}$, where  the 
closest one to $(g_4,g_5,g_6)$  implies $\mathfrak{O}$ (see Figs.\ \ref{fig:2} and \ref{fig:3}).

\paragraph{Fixed Position:}
Now we ask for the closest singular configuration $\mathfrak{P}$, which has the same position $(g_4,g_5,g_6)$ as the given pose $\mathfrak{G}$. 
As $\mathfrak{G}$ and  $\mathfrak{P}$ only differ in orientation, the angle $\in[0,\pi]$ enclosed by these two directions 
can be used as distance function. Note that this angle is the spherical distance function on $S^2$. 

By intersecting the singularity surface for the given position $(g_4,g_5,g_6)$ with $S^2$ we obtain a spherical curve $\omega(g_4,g_5,g_6)$ of degree 4. 
Then $\mathfrak{P}$ has to be a spherical pedal-point on $\omega(g_4,g_5,g_6)$ with 
respect to the point $(g_1,g_2,g_3)\in S^2$ (see Fig.\ \ref{fig:2}). 
By replacing the underlying spherical distance by the Euclidean metric of the ambient space $\mathbb{R}^3$, 
one will not change the set $\mathcal{P}$ of pedal-points on $\omega(g_4,g_5,g_6)$ with respect to $(g_1,g_2,g_3)$. 
Therefore $\mathcal{P}$ can be computed as the variety 
$V(\tfrac{\partial L}{\partial u}, \tfrac{\partial L}{\partial v}, \tfrac{\partial L}{\partial w}, 
\tfrac{\partial L}{\partial\lambda_{1}}, \tfrac{\partial L}{\partial\lambda_{2}})$
where $\lambda_1$ and $\lambda_2$ are the Lagrange multipliers of the Lagrange equation 
\begin{equation}
L(u,v,w,\lambda_{1},\lambda_{2})=(u-g_{1})^{2}+(v-g_{2})^{2}+(w-g_{3})^{2}+\lambda_{1}F+\lambda_{2}G
\end{equation}
with $G=u^{2}+v^{2}+w^{2}-1$. 
It can easily be checked (see Appendix B) that in general $\mathcal{P}$ consists of 8 points over $\mathbb{C}$, where the one with  
the shortest spherical distance to $(g_4,g_5,g_6)$ implies  $\mathfrak{P}$ (see Fig.\ \ref{fig:3}). 

\begin{remark}
For the practical application of this spherical distance to the singularity, we recommend to locate the position vector $\mathbf{p}$ 
in the tool-center-point of $\ell$. \hfill $\diamond$ 
\end{remark}

\newpage
\paragraph{General Case:} 
In contrast to the two special cases discussed above, the general case deals with mixed (translational and rotational) DOFs, thus 
the question of a suitable distance function arises.  
As the configuration space $\mathcal{C}$ equals the space of oriented line-elements, 
we can adopt the object dependent metrics discussed in \cite{naccepted} for our 
mechanical device as follows:
\begin{equation}\label{distance}
d(\mathfrak{L}, \mathfrak{L^{'}})^{2}:=
\frac{1}{5}\sum_{j=1}^5{\|\mathbf{b}_j-\mathbf{b}^{'}_{j}\|}^{2}
\end{equation}
where $\mathfrak{L}$ and $\mathfrak{L^{'}}$ are two configurations and $\mathbf{b}_j$ and $\mathbf{b}^{'}_{j}$ denote
the coordinate vectors of the corresponding platform anchor points. 
Note that the ambient space $\mathbb{R}^6$ (of $\mathcal{C}$) equipped with the metric $d$ of Eq.\ (\ref{distance}) is a 
Euclidean space (cf.\ \cite{naccepted}). 

With respect to this metric $d$ we can compute the closest singular configuration $\mathfrak{M}$ to $\mathfrak{G}$ in the following way: 
We determine the set $\mathcal{M}$ of pedal-points on the singularity variety with respect to $\mathfrak{G}$ as the variety 
$V(\tfrac{\partial L}{\partial u}, \tfrac{\partial L}{\partial v}, \tfrac{\partial L}{\partial w}, \tfrac{\partial L}{\partial p_{x}}, 
\tfrac{\partial L}{\partial p_{y}}, \tfrac{\partial L}{\partial p_{z}}, \tfrac{\partial L}{\partial \lambda_{1}}, 
\tfrac{\partial L}{\partial \lambda_{2}})$
where $\lambda_1$ and $\lambda_2$ are the Lagrange multipliers of the Lagrange equation 
\begin{equation}
L(u,v,w,p_{x},p_{y}, p_{z}, \lambda_{1}, \lambda_{2}):=d(\mathfrak{M}, \mathfrak{G})^{2} + \lambda_{1}G + \lambda_{2}F.
\end{equation} 
Random examples (see Appendix C) indicate that $\mathcal{M}$ consists of eighty points over $\mathbb{C}$, where the one with  
the shortest distance $d$ to $\mathfrak{G}$ equals $\mathfrak{M}$ (see Fig.\ \ref{fig:3}).

\begin{remark}
Note that these minimal distances can be seen as the radii of maximal singularity-free 
hyperspheres \cite{li2007determination} in the position workspace (see also \cite{nag2016}), 
the orientation workspace (see also \cite{jg2009}) and the complete configuration space. 
Moreover the distance $d(\mathfrak{M}, \mathfrak{G})$ to the singularity variety can also be interpreted as 
quality index thus it is an alternative to the value of $F$ proposed in \cite{btoc2009}. \hfill $\diamond$
\end{remark}

\newpage

\section{Conclusions and future research}

We presented a rational parametrization of the singularity variety of linear pentapods 
in Section \ref{sec:rat} and computed the distance to it in Section \ref{sec:dis}  
with respect to the novel metric given in  Eq. (\ref{distance}), which can easily be adopted 
for e.g.\ Stewart PMs as well. As this distance is of interest for many tasks 
(e.g.\ quality index for path planning, radius of the maximal singularity-free hypersphere,  \ldots) 
a detailed study of it (e.g.\ efficient computation of $\mathfrak{M}$, proof of $\#\mathcal{M}=80$, \ldots ) is dedicated to future research.

\begin{acknowledgement}
The first author is funded by the Doctoral College \emph{Computational Design}
of the Vienna University of Technology. The second author is supported by
Grant No. P 24927-N25 of the Austrian Science Fund FWF within the project
\emph{Stewart Gough platforms with self-motions}.
\end{acknowledgement}

\newpage

\subsection*{Appendix A}


The set $\mathcal{O}$ equals the variety of the ideal 
$\langle
\tfrac{\partial L}{\partial p_{x}}, \tfrac{\partial L}{\partial p_{y}}, \tfrac{\partial L}{\partial p_{z}}, \tfrac{\partial L}{\partial \lambda} 
\rangle$,
which can be computed as follows:

As we are dealing with a fixed orientation we can assume without loss of generality that $r_1=0$ holds beside the conditions given in 
Eq.\ (\ref{assumptions}). It turns out that the equations 
$\tfrac{\partial L}{\partial p_{x}}= \tfrac{\partial L}{\partial p_{y}}= \tfrac{\partial L}{\partial p_{z}}=0$ 
are linear with respect to $p_x,p_{y},p_z$. By solving these equations 
for the variables $p_x,p_{y},p_z$ and by plugging the obtained expressions into $\tfrac{\partial L}{\partial \lambda}$ it is shown that the
numerator $K$, which has 354513 terms, is of degree $6$ in $\lambda$. 


\paragraph*{Solutions:}
It turns out that for the random example (i.e.\
architectural parameters and pose $\mathfrak{G}$ given in the captions of Figs.\ \ref{fig:1} and \ref{fig:2}) 
the equation $K=0$ has $4$ real solutions and $2$ complex ones. The corresponding values of  $p_x,p_{y},p_z$ 
are obtained by back-substitution \vspace{-5mm} (cf.\ Table \ref{table:values1}).

\begin{small}
\begin{table}[h]
\centering
\begin{tabular}{|c|c|c|c|c|c|}
\hline 
\phantom{1} & $p_{x}$ & $p_{y}$ & $p_{z}$ & $\lambda$ & $l$  \\ \hline \hline 
1 & 2.551763090 & 2.637467970 & 0.1144666998 & 0.0002811301 & 3.941223289   \\ \hline
2 & 0.4205946500 & -10.11287492 & 3.678294530 & 0.0045791513  & 13.21156707   \\ \hline
3 & -6.106365796 & -8.333480392 & -0.3367715809 & 0.7825158446 & 14.30080937   \\ \hline
4 & -39.77559922 & -14.40064789 & -6.535304462 & -0.6930082534 & 46.46478104   \\ \hline
\end{tabular}
\caption{The 4 real solutions in ascending order with respect to the length $l$ of the translation 
vector towards the given \vspace{-5mm}position. 
}
\label{table:values1}
\end{table}
\end{small}

\subsection*{Appendix B}

The set $\mathcal{P}$ equals the variety of the ideal 
$\langle
\tfrac{\partial L}{\partial u}, \tfrac{\partial L}{\partial v}, \tfrac{\partial L}{\partial w}, \tfrac{\partial L}{\partial \lambda_{1}}, \tfrac{\partial L}{\partial \lambda_{2}} 
\rangle$, 
which can be computed as follows:

Under consideration of our assumptions given in Eq.\ (\ref{assumptions}), 
we start computing $\lambda_{1}, \lambda_{2}$ from the two equations 
$\tfrac{\partial L}{\partial u}= \tfrac{\partial L}{\partial v}=0$, 
which are linear in $u,v$ and $w$. By plugging the obtained expressions into $\tfrac{\partial L}{\partial w}$, $\tfrac{\partial L}{\partial \lambda_{1}}$ and $\tfrac{\partial L}{\partial \lambda_{2}}$ we get three rational polynomials in the variables $u$, $v$ and $w$. 
We name their numerators $F_{1}$, $F_{2}$ and $F_{3}$, respectively. It turns out that these equations are quadratic. Since the solution set of these quadratic equations is $V(F_{1}, F_{2}, F_{3})=V(F_{1})\cap V(F_{2})\cap V(F_{3})$, the number of solutions is $8$ according to \emph{Bezout's Theorem}. Additionally the number of terms of the polynomials $F_{1}$, $F_{2}$ and $F_{3}$ are $896$, $348$ and $4$ respectively ($F_{3}$ is in fact the implicit equation of the sphere).   
Now in order to obtain these 8 solutions we use the resultant method in the following form:
\begin{equation}
R_1:=Res(F_{2},F_{3},u), \quad
R_2:=Res(F_{1},F_{3},u), \quad
R_3:=Res(F_{1},F_{2},u),
\end{equation}
where $R_1, R_2$ and $R_3$ are dependent on the variables $v$ and $w$.
By using the resultant method again to eliminate the variable $v$ we obtain
\begin{equation}
G_1:=Res(R_{2},R_{3},v), \quad
G_2:=Res(R_{1},R_{3},v), \quad
G_3:=Res(R_{1},R_{2},v). 
\end{equation}
The greatest common divisor of $G_1, G_2, G_3$ yields the degree 8 polynomial in $w$. 

\paragraph*{Solutions:}
It turns out that for the random example (i.e.\
architectural parameters and pose $\mathfrak{G}$ given in the captions of Figs.\ \ref{fig:1} and \ref{fig:2}) 
only $2$ solutions are real. The corresponding values of  $u,v,w,\lambda_2$ 
are obtained by back-substitution \vspace{-5mm} (cf.\ Table \ref{table:values2}).


\begin{small}
\begin{table}[h]
\centering
\begin{tabular}{|c|c|c|c|c|c|c|}
\hline 
\phantom{1} & $u$ & $v$ & $w$ & $\lambda_{1}$ & $\lambda_{2}$ & $s$  \\ \hline \hline 
1 & 0.3701933149 & 0.5523718708 & 0.7468883632 & $0.0000002748$ & 4.381351180 & $48.41786560^{\circ}$   \\ \hline
2 & -0.3265950579 & -0.5850572044 & -0.7423232012 & 0.0000131667 & -6.434527722 & $131.6726142^{\circ}$   \\ \hline
\end{tabular}
\caption{The 2 real solutions in ascending order with respect to the spherical distance $s$ 
to the given \vspace{-5mm} orientation. 
}
\label{table:values2}
\end{table}
\end{small}
\subsection*{Appendix C}
\paragraph*{Gr\"obner Base:}

It is possible to compute the Gr\"obner basis of the ideal 
\begin{equation}
\langle
\tfrac{\partial L}{\partial u}, \tfrac{\partial L}{\partial v}, \tfrac{\partial L}{\partial w}, \tfrac{\partial L}{\partial p_{x}}, 
\tfrac{\partial L}{\partial p_{y}}, \tfrac{\partial L}{\partial p_{z}}, \tfrac{\partial L}{\partial \lambda_{1}}, 
\tfrac{\partial L}{\partial \lambda_{2}}
\rangle
\end{equation}
by {\tt Maple} using the FGb package of Faug\`{e}re \cite{faugere} for a random example (e.g.\  architectural parameters and 
pose $\mathfrak{G}$ given in the captions of Figs.\ \ref{fig:1} and \ref{fig:2}).
By means of this package we can also compute the univariate polynomial $P$ in $u$. 
The corresponding {\tt Maple} pseudo-code reads as follows: 
\begin{equation*}
\begin{split}
&with(FGb); \\
&GB:=\text{fgb\_gbasis}([\tfrac{\partial L}{\partial u}, \tfrac{\partial L}{\partial v}, \tfrac{\partial L}{\partial w}, \tfrac{\partial L}{\partial p_{x}}, 
\tfrac{\partial L}{\partial p_{y}}, \tfrac{\partial L}{\partial p_{z}}, \tfrac{\partial L}{\partial \lambda_{1}}, 
\tfrac{\partial L}{\partial \lambda_{2}}],0,[\,],[u,v,w,\lambda_2,\lambda_1,p_x,p_y,p_z]): \\
&P:=\text{fgb\_gbasis\_elim}(GB,0,[v,w,\lambda_2,\lambda_1,p_x,p_y,p_z],[u,v,w,\lambda_2,\lambda_1,p_x,p_y,p_z]):
\end{split}
\end{equation*}

\noindent
It can easily be checked that $P$ is of degree $80$ in $u$. 

\paragraph*{Resultant method:}
We are also able to compute this polynomial $P$ by a 
stepwise elimination of unknowns based on resultant method executed by {\tt Maple}. Details of this approach read as 
follows\footnote{Degrees and lengths of the given polynomials and factors are given with respect to the 
architectural parameters and pose $\mathfrak{G}$ given in the captions of Figs.\ \ref{fig:1} and \ref{fig:2}, respectively.}:
We start by computing $p_x,p_{y},p_z$ from the three equations 
$\tfrac{\partial L}{\partial p_{x}}= \tfrac{\partial L}{\partial p_{y}}= \tfrac{\partial L}{\partial p_{z}}=0$, 
which are linear in $p_x,p_y,p_z$. Plugging the obtained expressions into $\tfrac{\partial L}{\partial u}$ shows that its 
numerator only depends linearly on $\lambda_1$. From this condition we compute $\lambda_1$ and insert it into the 
equations $\tfrac{\partial L}{\partial v}= \tfrac{\partial L}{\partial w}= 
\tfrac{\partial L}{\partial \lambda_{2}}=0$, which only depend on $u,v,w,\lambda_2$. 
The remaining equation $\tfrac{\partial L}{\partial \lambda_{1}}= 0$ equals $G=0$ with $G=u^{2}+v^{2}+w^{2}-1$.
Then we compute the following resultants: 
\begin{equation*}
H_1:=Res(G_v,G,w), \quad
H_2:=Res(G_w,G,w), \quad
H_3:=Res(G_{\lambda_2},G,w), 
\end{equation*}
where $G_i$ with $i\in\left\{v,w,\lambda_2\right\}$ denotes the numerator of $\tfrac{\partial L}{\partial i}(u,v,w,\lambda_2)$. 
Note that $G_v$ is of degree $8$ in $w$ and that $G_w$ and  $G_{\lambda_2}$ are both of degree $9$ in $w$. Moreover we have 
\begin{equation*}
H_1[1230], \quad
H_2[1271], \quad
H_3[1252], 
\end{equation*}
where the number in the brackets gives the number of terms. It should also be mentioned that $H_1$ and $H_2$ are polynomials of degree $14$ with 
respect to $\lambda_2$ and that $H_3$ is of degree $12$ in $\lambda_2$. 
Then we proceed by computing
\begin{equation*}
K_1:=Res(H_2,H_3,\lambda_2), \quad
K_2:=Res(H_1,H_3,\lambda_2), \quad
K_3:=Res(H_1,H_2,\lambda_2).  
\end{equation*}
$K_1,K_2,K_3$ have two common factors, which do not cause solutions as they imply zeros in the denominators of above arisen expressions. 
Beside these factors $K_1$, $K_2$, $K_3$ split up into 
\begin{equation*}
K_{1,1}[2016]K_{1,2}[11175], \quad
K_{2,1}[1938]K_{2,2}[11097] \quad \text{and} \quad
K_{3,1}[1653]K_{3,2}[11371],
\end{equation*}
respectively, where the long factors $K_{j,2}$ (for $j=1,2,3$)  are caused by the elimination process and do not 
contribute to the final solution. The factors $K_{1,1}$ and $K_{2,1}$ are of degree $62$ in $v$ and $K_{3,1}$ is of degree 
$56$ in $v$. The greatest common divisor of $Res(K_{1,1},K_{3,1},v)$ and $Res(K_{2,1},K_{3,1},v)$ yields the univariate polynomial $P$ in $u$.  

\paragraph*{Solutions:}
The polynomial $P$ (either obtained by  Gr\"obner basis elimination techniques or by the resultant method) has to be solved 
numerically. It turns out that for the random example under consideration only $16$ solutions are real and $64$ solutions are 
complex.\footnote{It is unknown if examples with $80$ real solutions can exist.}  
By back-substitution into the equations obtained during the stepwise elimination based on resultant method, we get the 
values for \vspace{-5mm} $u,v,w,\lambda_2$ (cf.\ Table \ref{table:values}). 

\begin{small}
\begin{table}[h]
\centering
\begin{tabular}{|c|c|c|c|c|c|}
\hline 
\phantom{1} & $u$ & $v$ & $w$ & $\lambda_2$ & $d$  \\ \hline \hline 
1 & 0.5559273038 & 0.7274604486 & 0.4021767380 & 0.0000977412 & 1.479192394   \\ \hline
2 & 0.7100848787 & 0.6097073464 & -0.3521880419 & 0.0112120286 & 6.370089783   \\ \hline
3 & 0.6707364219 & 0.6608309577 & -0.3367715809 & 0.0263760716 & 6.396348687   \\ \hline
4 & 0.9520812787 & 0.2971145357 & -0.0725547483 & 0.0518935724 & 6.494930694   \\ \hline
5 & -0.4198912232 & -0.7478308408 & -0.5142376826 & 0.0001295064 & 6.522840484   \\ \hline
6 & 0.6426323048 & 0.5670451826 & -0.5152508920 & 0.6390457179 & 7.901089998   \\ \hline
7 & -0.9141441020 & 0.2145188032 & -0.3439800050 & 0.0175335168 & 8.153560918   \\ \hline
8 & -0.6633066166 & -0.6523793948 & 0.3666407746 & -0.0198075803 & 9.072642063   \\ \hline
9 & -0.4968498376 & -0.8534603513 & -0.1573075574 & 0.0025204733 & 9.244102979   \\ \hline
10 & 0.4561177696 & -0.7759229223 & 0.4357753997 & -0.1291505456 & 9.308167139   \\ \hline
11 & -0.6449198390 & -0.5507646090 & 0.5298459651 & 0.0516473389 & 9.970322913   \\ \hline
12 & 0.9794782799 & -0.1982747821 & 0.0361857697 & 0.0967973416 & 10.05488078   \\ \hline
13 & -0.2161351180 & 0.9213140919 & -0.3232119349 & 0.0362890328 & 13.78049458   \\ \hline
14 & 0.1003162322 & -0.5648716701 & -0.8190583922 & 0.0430362461 & 37.60374403   \\ \hline
15 & 0.8001243699 & 0.1449203823 & 0.5820644943 & -0.1991801120 & 52.29308488   \\ \hline
16 & 0.0428010579 & 0.5580832041 & 0.8286804008 & -0.0557997298 & 65.26242524   \\ \hline
\end{tabular}
\caption{The 16 real solutions in ascending order with respect to the distance $d$ (given in 
Eq. (\ref{distance})) from $\mathfrak{G}$. The corresponding values of missing variables $p_x,p_y,p_z,\lambda_1$ 
are obtained by substituting $u,v,w,\lambda_2$  into the expressions for $p_x,p_y,p_z,\lambda_1$. 
For the global minimizer (solution 1) these values are 
$p_x=2.296688437$, 
$p_y=3.479406728$, 
$p_z=1.835729103$ and 
$\lambda_1=-4.720444174$.
}
\label{table:values}
\end{table}
\end{small}

\newpage

\section*{Addendum}

We can simplify the problem by considering equiform transformations of the 
linear platform $\ell$. This means that we can cancel the side condition 
$G=1$. The computation can be done in a similar fashion to Appendic C with the sole 
difference that we set $\lambda_1=0$. 

\begin{figure}[t!] 
\begin{center}   
  \begin{overpic}[height=51mm]{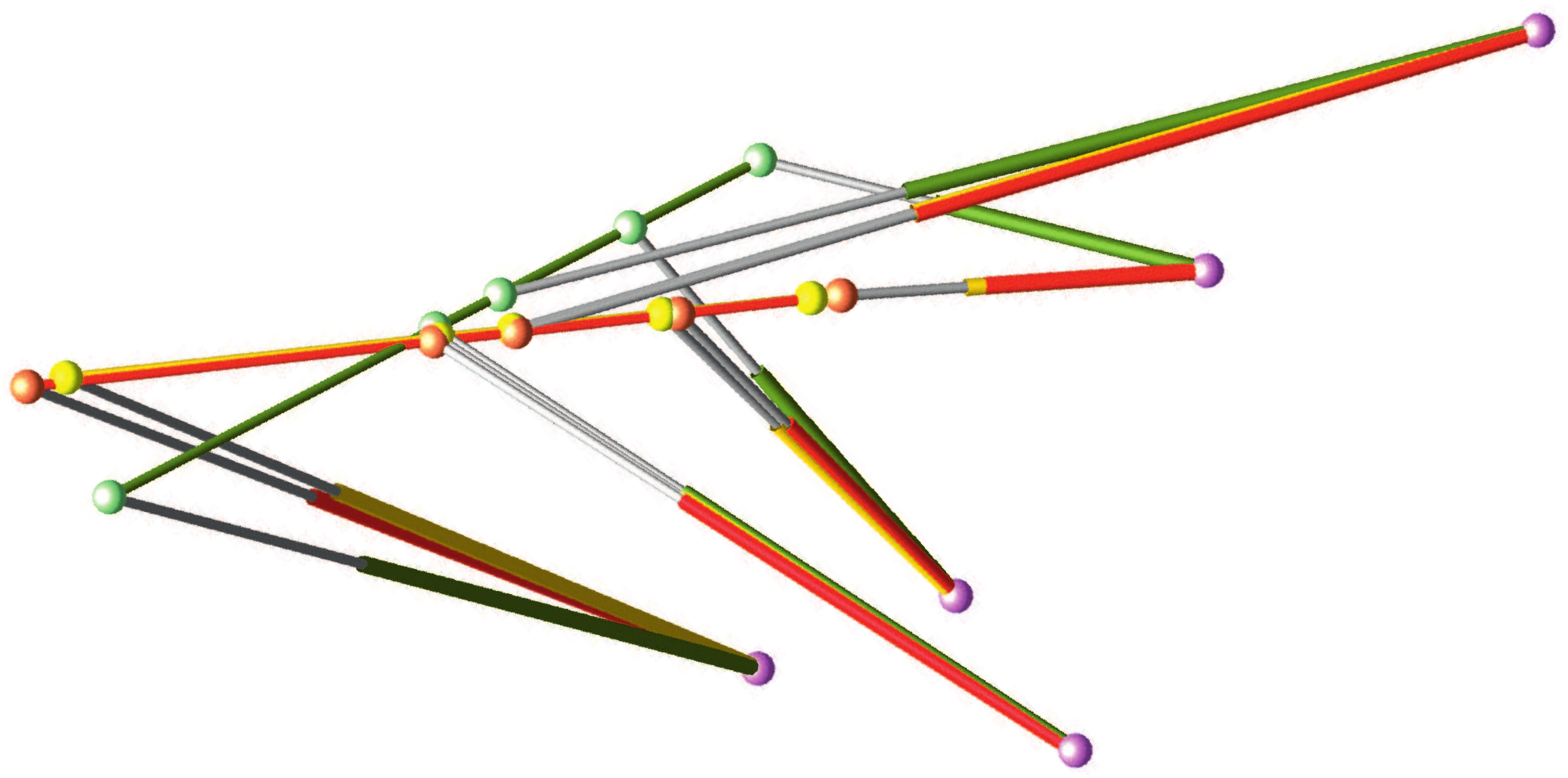}
  \end{overpic} 
	\caption{Comparison of the different configurations $\mathfrak{G}$ (green), 
		$\mathfrak{M}$ (red) and $\mathfrak{N}$ (yellow). Note that 
		$d(\mathfrak{N}, \mathfrak{G})=1.4517670618$ is smaller than  $d(\mathfrak{M}, \mathfrak{G})=1.479192394$. 
	}
	\label{fig:4}
\end{center}
\end{figure}

Random examples show that this reduced problem has only $28$ solutions over $\mathbb{C}$ in the 
general case. For the architectural parameters and pose $\mathfrak{G}$ given in the captions of Figs.\ 
\ref{fig:1} and \ref{fig:2} it turns out that only $6$ solutions are real, which are given in 
Table \ref{table:values3}. Moreover the global minimizer $\mathfrak{N}$ is displayed in Fig.\ \ref{fig:4}. 
Important for application is that $d(\mathfrak{N}, \mathfrak{G})\leq d(\mathfrak{M}, \mathfrak{G})$ and therefore 
the value of  $d(\mathfrak{N}, \mathfrak{G})$ gives us the radius of hyper-sphere, which is guaranteed  singularity free.

\begin{small}
\begin{table}[h]
\centering
\begin{tabular}{|c|c|c|c|c|c|c|}
\hline 
\phantom{1} & $u$ & $v$ & $w$ & $\lambda_2$ & $d$ & $\mu$  \\ \hline \hline 
1 & 0.5055836745 &  0.6656442614 &  0.3718172932 &  0.0000990198 & 1.4517670618 & 0.9148471097  \\ \hline
2 & 0.6486166479 &  0.5542384068 & -0.3254994305 &  0.0118010768 & 6.3636100364 & 0.9131449210  \\ \hline
3 & 0.6200661016 &  0.6110910452 & -0.3126505854 &  0.0283247861 & 6.3914193483 & 0.9250214183  \\ \hline
4 & 0.8789381698 &  0.2541301769 & -0.0569840426 &  0.0548756462 & 6.4897306508 & 0.9167124054  \\ \hline
5 & 0.5220226650 &  0.4591892110 & -0.4197574551 &  9.9482986641 & 7.8756112220 & 0.8121322031  \\ \hline
6 & 0.6513903749 & -0.9060852905 &  0.5534569690 & -0.1501405543 & 9.2038614723 & 1.2456382263  \\ \hline
\end{tabular}
\caption{The 6 real solutions in ascending order with respect to the distance $d$ (given in 
Eq. (\ref{distance})) from $\mathfrak{G}$. The scaling factor of the corresponding 
equiform displacement of the platform is given by $\mu$.
The corresponding values of missing variables $p_x,p_y,p_z$ 
are obtained by substituting $u,v,w,\lambda_2$  into the expressions for $p_x,p_y,p_z$. 
For the global minimizer $\mathfrak{N}$ (solution 1) these values are 
$p_x=2.5031164070$, 
$p_y=3.7266491740$ and 
$p_z=1.9989579769$. 
}
\label{table:values3}
\end{table}
\end{small}

\end{document}